\documentclass[letterpaper, 10 pt, conference]{ieeeconf}

\IEEEoverridecommandlockouts %
\usepackage{times}

\usepackage{amsmath,amssymb}
\usepackage{amsfonts}
\usepackage{graphicx}
\usepackage{textcomp}
\usepackage{xcolor}
\usepackage{soul}
\usepackage{titletoc}
\usepackage{float}
\usepackage{listings}
\usepackage{multirow}
\usepackage{xparse}
\usepackage{optidef}
\usepackage{algorithm}
\usepackage{algpseudocode}
\usepackage{wrapfig}
\usepackage{url}
\usepackage{tikz}
\usetikzlibrary{arrows.meta,positioning,fit,backgrounds,calc}

\usepackage{multicol}
\usepackage{booktabs}
\usepackage{bm}
\usepackage{overpic}

\usepackage[normalem]{ulem} %

\definecolor{code-constant}{HTML}{d86001}
\definecolor{code-grey}{HTML}{848482} %

\usepackage[noadjust]{cite}

\usepackage{color}
\definecolor{international_orange}{RGB}{240, 74, 0}
\definecolor{citecolor}{RGB}{240, 74, 0}
\usepackage[pagebackref=true,breaklinks=true,urlcolor=citecolor,colorlinks,citecolor=citecolor,bookmarks=false,linkcolor=citecolor]{hyperref}

\usepackage{cleveref}

\newcommand{\datafarm}{DATAFARM}
\newcommand{\algname}{\underline{D}istribution-\underline{A}ligned \underline{T}ask \underline{A}nd motion planning for \underline{F}ine-tuning \underline{A} \underline{R}obot foundation \underline{M}odel}

\newcommand{\basepretrained}{$\pi_{0.5}$-DROID}
\newcommand{\basetamp}{Raw TAMP}
\newcommand{\baseteleop}{Human teleop (oracle)}
\newcommand{\ours}{\datafarm{}}
\newcommand{\ablstyle}{\datafarm{} w/o style}
\newcommand{\abltiming}{\datafarm{} w/o timing}
\newcommand{\abljoint}{\datafarm{} w/o joint space}
\newcommand{\taskplate}{Geometric Constraints}
\newcommand{\taskpack}{Multi-Step Reasoning}
\newcommand{\tasksort}{Semantic Reasoning}
\newcommand{\taskcloth}{Deformable Object Manipulation}
\newcommand{\evalmain}{Target}
\newcommand{\evalood}{OOD}
\newcommand{\pretrainpolicy}{\basepretrained{}}
\newcommand{\best}[1]{\textbf{#1}}

\NewDocumentCommand \proposition {g g g g} {\texttt{#1}(#2
  \IfValueTF{#3}{,\,#3}{}
  \IfValueTF{#4}{,\,#4}{}
  )
}

\definecolor{revision_blue}{RGB}{0, 0, 200}

\usepackage{subcaption}
\usepackage[skip=0pt,font=small,labelfont=bf]{caption}
\usepackage{amsmath} 
\newcommand{\shrinka}

\title{\textbf{\datafarm{}}: Distribution-Aligned Task and Motion Planning \\ for Fine-Tuning Vision-Language-Action Models}

\begin{document}
\author{%
  Samrat Sahoo$^{1}$,
  Yixuan Huang$^{2}$, 
  and Tom Silver$^{2}$ 
  \thanks{$^{1}$%
    Stanford University.
    $^{2}$ Princeton University. Work done while Samrat was visiting Princeton. samrat@stanford.edu, yixuan.huang@princeton.edu, and tsilver@princeton.edu. 
  }%
}

\input{overview_fig}
\maketitle

\begin{abstract}

Collecting high-quality robot data remains a fundamental challenge for training robot foundation models. 
Task and motion planning~(TAMP) offers a scalable way to generate demonstrations, but our experiments show that raw TAMP trajectories provide surprisingly little benefit when used to fine-tune pretrained vision-language-action (VLA) models, despite successfully solving the target tasks. 
We hypothesize that this failure arises from a behavioral distribution mismatch between planner-generated trajectories and the data used to pretrain the VLA.
To address this mismatch, we introduce \ours: \algname{}, an approach that incorporates the pretraining distribution directly into TAMP trajectory generation. 
\ours{} aligns generated trajectories with the pretraining data in robot joint configurations, motion style, and temporal execution profiles. 
We evaluate \ours{} on three tabletop manipulation tasks that TAMP can perform and a cloth-folding task beyond the capability of TAMP.
\ours{} achieves an average success rate of 56.7\%, substantially outperforming raw TAMP~(8.3\%) while approaching human teleoperation~(61.7\%). 
On \taskcloth{}, which is outside the fine-tuning distribution, the fine-tuned model retains 85\% success, compared with 90\% for the pretrained model.
These results show that aligning planner-generated demonstrations with the pretraining distribution can make TAMP an effective source of data for VLA fine-tuning. Website
and code: \url{https://prpl-group.com/datafarm/}

\end{abstract}

\section{Introduction}\label{sec:introduction}
Task and motion planning (TAMP)~\cite{garrett2021integrated} can generate successful robot demonstrations for tasks that pretrained vision-language-action (VLA) models struggle to perform.
This creates an opportunity to extend these robot foundation models~\cite{black2025pi05, lbmtri2025} by fine-tuning them on planner-generated data.
The two approaches cover overlapping but distinct sets of tasks: TAMP is well suited to long-horizon manipulation with explicit geometric constraints, while the VLA can perform tasks beyond TAMP's scope, such as deformable object manipulation.
Our goal is to use TAMP demonstrations to teach the VLA additional tasks while preserving its existing capabilities.
For tasks within the planner's scope, this would allow us to collect fine-tuning data without a human operator.

Prior work has used TAMP-generated demonstrations to train visuomotor policies~\cite{dalal2023optimus, mandlekar2023hitltamp}.
We find, however, that successful planner trajectories provide little benefit when used directly to fine-tune a pretrained VLA.
Across three tabletop tasks, fine-tuning the pretrained VLA \basepretrained{} on demonstrations from TiPToP~\cite{shen2026tiptop}, which combines vision foundation models for perception with TAMP, raises average success from 1.7\% to only 8.3\%, even though TiPToP solves the tasks.
We hypothesize that this poor transfer reflects a mismatch between planner trajectories and pretraining data.
A task-valid trajectory can differ from pretraining trajectories in its joint configurations, motion style, or execution speed.

To address this behavioral mismatch, we introduce \datafarm{}, \algname{} (Figure~1).
\datafarm{} generates demonstrations for VLA fine-tuning by incorporating the pretraining distribution into TAMP's trajectory optimization objectives.
We first train an encoder on pretraining trajectories and fit a distribution over their learned representations.
Our experiments use DROID~\cite{khazatsky2024droid}, a dataset of robot manipulation demonstrations collected through human teleoperation, which was used to train the VLA checkpoint.
During planning, the frozen encoder maps candidate trajectories into this representation space, where their distance from the fitted distribution provides an objective for optimizing motion style and execution timing.
A separate model of joint configurations guides motion planning endpoint selection.
Throughout this process, the planner enforces the task and motion constraints required for a valid solution.

We integrate \datafarm{} into TiPToP and use its demonstrations to fine-tune \basepretrained{} on three tabletop tasks.
Average success is \textbf{56.7\%} for \textbf{\datafarm{}}, compared with \textbf{8.3\%} for \textbf{raw TAMP} and \textbf{61.7\%} for \textbf{human teleoperation}.
Ablations show that all three alignment objectives contribute to the improvement.
We also test whether fine-tuning preserves the model's performance on cloth folding, an instance of deformable object manipulation, which is beyond our planner's scope.
The fine-tuned VLA achieves 85\% success on cloth folding, compared with 90\% for the pretrained model, without using demonstrations of this task for fine-tuning.
These results show that distribution alignment helps the VLA learn from the planner while largely preserving its ability to perform a task the planner cannot.

\section{Related Work}\label{sec:related-work}

\subsection{Planning-Based Demonstration Generation}

Planning can reduce the cost of policy learning by providing demonstrations that would otherwise require human operators. Prior work has used TAMP to train visuomotor policies~\cite{mcdonald2022guided,dalal2023optimus}, combined language-guided task decomposition with motion planning to train multitask policies~\cite{ha2023scaling}, and generated demonstrations for contact-rich manipulation using model-based planners~\cite{li2024glide}. Zhu et al.\ also examine the suitability of planner-generated demonstrations for policy learning, showing that improving demonstration consistency benefits behavior cloning~\cite{zhu2025planners}. Other approaches combine planning with learned manipulation skills. HITL-TAMP uses a planner to automate portions of demonstration collection and learns policies for the remaining human-controlled segments~\cite{mandlekar2023hitltamp}; SkillMimicGen adapts human skill demonstrations and connects them through planned motions~\cite{garrett2025skillmimicgen}. Both retain planning as part of the deployed controller. DATAFARM uses planner-generated fine-tuning data to teach new tasks to an already pretrained VLA while retaining its existing capabilities, including behaviors beyond the planner's scope.

\subsection{Demonstration Collection and Augmentation}

Large teleoperation datasets such as DROID~\cite{khazatsky2024droid} and portable collection interfaces such as UMI~\cite{chi2024umi} broaden the range of demonstrations available for robot learning, but collection remains limited by human time. Demonstration augmentation amortizes this effort by generating additional data from a small number of examples. MimicGen and DemoGen adapt demonstrations to new object configurations~\cite{mandlekar2023mimicgen,xue2025demogen}. CP-Gen combines demonstration-derived constraints with optimization and motion planning to accommodate variations in object geometry and pose~\cite{lin2025cpgen}, while physics-driven generation combines demonstration retargeting and trajectory optimization for contact-rich manipulation~\cite{yang2025physicsgen}. DATAFARM also reuses human data, but does not require examples of the target tasks. Here, pretraining demonstrations supply a prior over robot behavior, while the planner supplies solutions to the new tasks.

\subsection{Synthetic Data for VLA Fine-Tuning}

Recent approaches generate synthetic data for adapting pretrained VLAs such as $\pi_{0.5}$~\cite{black2025pi05}. AOMGen and Pose6DAug expand demonstrations through changes to objects and their visual context while maintaining consistency with the underlying interaction~\cite{wu2025aomgen,lee2026pose6daug}. RoboPaint and Real2Render2Real convert human demonstrations into robot training data through retargeting and rendering~\cite{fan2026robopaint,yu2025real2render2real}. DreamGen uses video generation and inferred action labels to produce training data for VLAs~\cite{jang2025dreamgen}. In concurrent work, Sato et al.~\cite{sato2026dream} introduce DREAM, which combines workspace reconstruction with TAMP to generate demonstrations for VLA fine-tuning without task-specific human examples. DREAM focuses on generating data for the deployment workspace and does not align planner behavior with the VLA's pretraining distribution. DATAFARM incorporates this distribution into planning objectives for joint configurations, motion style, and execution timing to improve the usefulness of generated demonstrations for fine-tuning.

\section{Problem Setup}\label{sec:problem-setup}
\begin{figure*}[t]
    \centering
    \includegraphics[width=\linewidth]{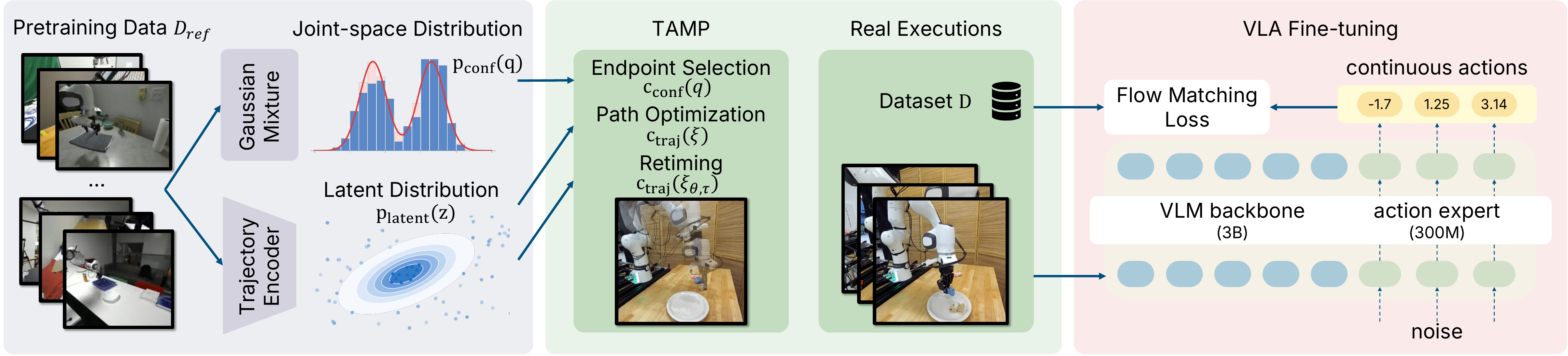}
    \hfill
    \caption{\textbf{Overview of \ours{}}. \ours{} first learns a reference distribution from pretraining demonstrations $D_{ref}$ by fitting a Gaussian mixture density $p_{conf}(q)$ and learning a trajectory encoder $E_{\phi}$ with a corresponding latent distribution $p_{latent}(z)$. 
    These references provide alignment costs for TAMP with endpoint selection, path optimization, and retiming. We execute the resulting trajectories on the robot to collect demonstrations $D$, which we use to fine-tune the pretrained VLA \pretrainpolicy{}.}
    \label{fig:approach}
\end{figure*}
We consider a robot with observation space $\mathcal{O}$ and action space
$\mathcal{A}$. Observations include camera images and proprioception.
Let $\mathcal{H}$ denote the space of observation--action histories and
$\mathcal{L}$ the space of natural-language instructions. A
vision-language-action (VLA) policy is a mapping
\[
\pi : \mathcal{H} \times \mathcal{L} \longrightarrow \Delta(\mathcal{A}),
\]
where $\Delta(\mathcal{A})$ denotes the set of distributions
over actions.

We are given a pretrained policy \pretrainpolicy{} and a finite set of target
manipulation tasks $\mathcal{T}$. Each task $T \in \mathcal{T}$ specifies
a language instruction, a distribution over initial scenes, and a
success criterion. A demonstration consists of an observation--action
trajectory paired with the task instruction. We are also given a
reference dataset $\mathcal{D}_{\mathrm{ref}}$ from a data source used to
pretrain \pretrainpolicy{}. These reference demonstrations include robot joint
trajectories but need not cover the target tasks.

Our goal is to develop a data-collection procedure that generates a
fine-tuning dataset $\mathcal{D}$ with $N$ successful demonstrations per target
task, for a given demonstration budget $N$. The robot must generate and
execute the demonstration motions autonomously; human assistance with
environment setup and resets is permitted. Each episode is recorded using
the observation and action interface expected by the VLA. Let
$\pi_{\mathcal{D}}$ denote the policy obtained by applying a fixed
fine-tuning procedure to \pretrainpolicy{} using $\mathcal{D}$.

We evaluate a policy $\pi$ by its expected success rate across
the target tasks:
\begin{equation}
J(\pi)
= \frac{1}{|\mathcal{T}|}
\sum_{T\in\mathcal{T}}
\Pr[\pi \text{ succeeds on }T].
\label{eq:datafarm-objective}
\end{equation}
Here, $J(\pi) \in [0,1]$, and the probability accounts for the task's
distribution over initial scenes and any stochasticity in policy
execution and the environment. We seek a
data-collection procedure that maximizes $J(\pi_{\mathcal{D}})$
under the demonstration budget, with evaluation performed on fresh
episodes. We also evaluate whether $\pi_{\mathcal{D}}$ retains the
pretrained policy's capabilities on tasks outside $\mathcal{T}$.

\section{Distribution-Aligned TAMP}\label{sec:approach}
\ours{} augments TAMP with objectives derived from the reference
demonstrations $\mathcal{D}_{\mathrm{ref}}$ to generate fine-tuning data
for $\pi_{0.5}$. 
We provide an overview of \ours{} in Figure~\ref{fig:approach}. 
We first review TiPToP's planning pipeline, then describe
how these objectives are constructed and incorporated into the planner. 

\subsection{Preliminaries}

Task and motion planning (TAMP) combines discrete decisions about which
actions to perform with continuous decisions about how to execute
them~\cite{garrett2021integrated}. For manipulation, these choices include
the order in which objects are moved, the grasps and placements used,
and the robot motions connecting them. These decisions are coupled:
a grasp that allows an object to be picked up may prevent it from being
placed at its destination. A solution must achieve the task goal while
satisfying geometric and kinematic constraints.

We build on TiPToP~\cite{shen2026tiptop}, a modular manipulation system
that constructs and executes TAMP solutions from images and language
instructions. Its perception module produces a 3D scene representation
with object geometries and candidate grasps, together with a symbolic
goal. Its planner, cuTAMP, enumerates candidate \emph{plan skeletons}:
sequences of symbolic actions whose continuous parameters remain
unspecified. A skeleton can, for example, move an obstructing object
before picking and placing the target object.

Let $q \in \mathcal{C}$ denote a robot configuration and
$x \in \mathrm{SE}(3)$ an end-effector pose, where $\mathcal{C}$ is the
robot's configuration space. For each skeleton, cuTAMP samples grasps
and placements and uses inverse kinematics (IK) to initialize robot
configurations. IK seeks configurations satisfying
$\operatorname{FK}(q)=x$, where $\operatorname{FK}$ is the
forward-kinematics map. Searching from multiple initial guesses can
produce a finite set of candidate solutions, denoted $\mathcal{Q}(x)$.
These candidates reach the same end-effector pose but can have different
arm postures. The planner must therefore choose robot configurations
in addition to grasps and placements. cuTAMP jointly refines placements
and configurations to satisfy collision-avoidance, placement-stability,
and kinematic constraints; the selected configurations become endpoints
for the motions between manipulation actions.

TiPToP invokes cuRobo~\cite{sundaralingam2023curobo} to generate these
connecting motions. We call each motion a \emph{trajectory segment} and
represent it by time-stamped waypoints
\[
\xi = \bigl((q_1,t_1),\ldots,(q_n,t_n)\bigr),
\]
where $q_i$ is the desired configuration at time $t_i$ and
$0=t_1<\cdots<t_n$. The first and last configurations, $q_1$ and $q_n$,
are the segment's \emph{endpoints}.

For fixed endpoints and sample times, trajectory optimization searches
over the intermediate waypoints to minimize an objective
$C_{\mathrm{plan}}(\xi)$. We write this optimization as
\begin{equation}
\begin{aligned}
\min_{q_{2:n-1}}\quad & C_{\mathrm{plan}}(\xi) \\
\text{with}\quad & q_1=q_{\mathrm{start}},\quad q_n=q_{\mathrm{goal}}.
\end{aligned}
\label{eq:base-trajopt}
\end{equation}
The objective combines smoothness costs, including penalties on
acceleration and jerk, with penalties for collisions and violations of
robot limits. cuRobo uses gradient-based optimization to iteratively
update the waypoints using derivatives of these costs with respect to
$q_{2:n-1}$. Motion derivatives are computed from the sampled
configurations and times using finite differences. Candidate trajectories
are checked for feasibility before being accepted.

A segment's \emph{geometric path} describes its motion through
configuration space, while its \emph{time parameterization} determines
how quickly that path is traversed. Even for a fixed path, changing the
traversal speed or duration changes the velocities, accelerations, and
jerks. We use \emph{retiming} to refer to changing the time
parameterization while keeping the geometric path fixed. cuRobo adjusts
its time discretization based on the robot's motion limits and reruns
trajectory optimization with the revised time step. TiPToP assembles
the segments with gripper commands into a full plan and tracks
the arm motion using a joint impedance controller.

\subsection{Method Overview}

DATAFARM learns models of reference behavior to guide the planner's
selection of configurations, paths, and execution times. From
$\mathcal{D}_{\mathrm{ref}}$, we fit a Gaussian mixture density
$p_{\mathrm{conf}}(q)$ over robot configurations. For trajectories, we
learn a differentiable encoder $E_{\phi}$ that maps a segment $\xi$ to a latent
vector $z=E_{\phi}(\xi)\in\mathbb{R}^{d}$, and fit a Gaussian density
$p_{\mathrm{latent}}(z)$ to the reference embeddings. These models are
learned before planning and subsequently held fixed.

The models provide two costs: a configuration cost
$c_{\mathrm{conf}}(q)=-\log p_{\mathrm{conf}}(q)$ and a trajectory cost
$c_{\mathrm{traj}}(\xi)$, defined by the squared Mahalanobis distance of
$E_{\phi}(\xi)$ from the reference latent distribution. The first evaluates
individual robot configurations; the second evaluates the motion over
an entire segment, including its temporal properties.

During planning, $c_{\mathrm{conf}}$ ranks the IK candidates in
$\mathcal{Q}(x)$. We add $c_{\mathrm{traj}}$ to the planner's objective
$C_{\mathrm{plan}}$ to optimize connecting paths with the selected
endpoints fixed, and use the same trajectory cost to guide retiming
along those paths. Gradients through the frozen encoder allow this cost
to influence the waypoint configurations and the time parameterization,
while the planner's feasibility checks remain in place. We execute the plans to collect demonstrations $\mathcal{D}$ for fine-tuning
$\pi_{0.5}$. The following subsections describe model learning, planning
with these costs, and data collection.

\subsection{Learning Reference Models}
\label{sec:reference-models}

\paragraph{Configuration density}
We fit $p_{\mathrm{conf}}$ as a Gaussian mixture with full-covariance
components to joint configurations in $\mathcal{D}_{\mathrm{ref}}$.
Only frames in which the arm is moving contribute to the fit, so that
long pauses do not concentrate the density on idle configurations.
The resulting cost $c_{\mathrm{conf}}(q)$ favors configurations that are
common during reference motions. It contains no information about
whether a configuration accomplishes a particular task; the planner
provides that constraint through IK and subsequent feasibility checks.

\paragraph{Trajectory representation}
To learn $E_{\phi}$, we extract arm trajectories from the reference episodes
and augment them with random temporal crops. Cropping exposes the
encoder to partial motions of different lengths, matching the segments
it will score during planning. Each trajectory is resampled at the
reference frequency $F$. At each sample, we concatenate joint positions
with their velocities, accelerations, and jerks, computed by finite
differences at spacing $1/F$. We standardize these channels using
reference-data statistics and keep the physical sampling interval fixed:
executing the same path at a different speed should change its
representation. Temporal pooling with padding masks allows the encoder
to accept variable-length inputs.

We train $E_{\phi}$ with an auxiliary prediction objective and variational
regularization~\cite{kingma2014autoencoding,higgins2017betavae}. During
training, the encoder also predicts a diagonal covariance, and we sample
$z$ from the Gaussian with mean $E_{\phi}(\xi)$ and this covariance. An
auxiliary head $g_{\psi}$ predicts a vector of motion descriptors
$f(\xi)$ from the sample. The descriptors summarize joint speeds,
accelerations, jerks, spectral content, path geometry, and smoothness.
We jointly optimize the encoder parameters $\phi$ and auxiliary-head
parameters $\psi$ using the expected weighted prediction error
\begin{equation}
\mathbb{E}_{z}\!\left[
\left\|w\odot\bigl(g_{\psi}(z)-f(\xi)\bigr)\right\|_2^2
\right],
\label{eq:descriptor-prediction}
\end{equation}
where the expectation is over the sampled latent vector and $w$ balances
the scales of the descriptors. We add a KL penalty from the predicted
Gaussian to $\mathcal{N}(0,I)$, weighted by $\beta$, and minimize the
combined objective over reference trajectories and their crops.
Predicting motion descriptors encourages the latent
representation to capture properties relevant to alignment without
requiring reconstruction of the full joint-position sequence. The
descriptors are training targets only: they are neither encoder inputs
nor evaluated during planning. The full training objective, architecture,
and descriptor definitions are given in Appendix.~\ref{app:latent}.

\paragraph{Latent reference density}
After training, we discard the auxiliary head and covariance output,
and freeze $\phi$. Planning uses the deterministic embedding $E_{\phi}(\xi)$, including
the resampling and standardization above. We fit
$p_{\mathrm{latent}}=\mathcal{N}(\mu,\Sigma)$ to embeddings of reference
trajectories and their crops, and define
\begin{equation}
c_{\mathrm{traj}}(\xi)
=\bigl(E_{\phi}(\xi)-\mu\bigr)^{\!\top}
\Sigma^{-1}\bigl(E_{\phi}(\xi)-\mu\bigr).
\label{eq:trajectory-cost}
\end{equation}
This cost measures deviation from the reference embeddings while
accounting for their covariance. The encoder, input normalization,
and both reference densities remain fixed throughout planning.

\subsection{Distribution-Aligned Planning}
\label{sec:aligned-planning}

\paragraph{Endpoint selection}
For each candidate end-effector pose $x$, we rank the IK solutions
in $\mathcal{Q}(x)$ using $c_{\mathrm{conf}}$ and select
\begin{equation}
q^{\star}
=\underset{q\in\mathcal{Q}(x)}{\operatorname{arg\,min}}
\;c_{\mathrm{conf}}(q).
\label{eq:endpoint-selection}
\end{equation}
This ranking favors arm postures with higher reference density among
the candidates reaching $x$. The selected configurations serve as
endpoints for path optimization, with grasps, placements, and connecting
motions subject to the planner's feasibility checks. Appendix.~\ref{app:joint} gives
the candidate-generation and validation details.

\paragraph{Path optimization}
With endpoints selected, we augment the objective in
Eq.~\eqref{eq:base-trajopt} with the trajectory cost:
\begin{equation}
\begin{aligned}
\min_{q_{2:n-1}}\quad
&C_{\mathrm{plan}}(\xi)
+\lambda_{\mathrm{path}}c_{\mathrm{traj}}(\xi)\\
\text{with}\quad
&q_1=q_{\mathrm{start}},\quad q_n=q_{\mathrm{goal}}.
\end{aligned}
\label{eq:aligned-trajopt}
\end{equation}
Here $\lambda_{\mathrm{path}}$ controls the contribution of the learned
cost, and the sample times remain fixed within each optimization pass.
The added cost can therefore change how the arm moves between the
endpoints, including the velocities and higher derivatives induced by
the waypoint sequence.

We optimize this objective using waypoint gradients, backpropagating
$c_{\mathrm{traj}}$ through the frozen encoder $E_{\phi}$ and the
kinematic preprocessing described above. The planner retains its
original constraint penalties and checks the resulting motion for
feasibility before accepting it. Appendix.~\ref{app:style} details the gradient
computation.

\paragraph{Retiming}
We next optimize how quickly the robot traverses each path produced by
the preceding optimization, keeping the path's geometry fixed.\footnote{%
In principle, path optimization and retiming could be performed jointly,
but we found that separate stages worked better in practice.}
Let $\gamma:[0,1]\to\mathcal C$ denote one such path, parameterized
by normalized arc length. A \emph{time scaling} is a nondecreasing
function $s:[0,\tau]\to[0,1]$ with $s(0)=0$ and $s(\tau)=1$; its
derivative $\dot s(t)$ is the speed profile in path coordinates.
The resulting configuration at time $t$ is $\gamma(s(t))$, with
joint velocity $\gamma'(s(t))\dot s(t)$~\cite{lynch2017modern}.

We parameterize the time scaling by dividing the path into $M$ equal
arc-length intervals and using $\theta\in\mathbb R^M$ to assign a
fraction $\alpha_m(\theta)$ of the total duration $\tau>0$ to each interval:
\begin{equation}
\Delta t_m=\tau\alpha_m(\theta),
\qquad m=1,\ldots,M.
\label{eq:time-parameterization}
\end{equation}
The fractions vary smoothly with $\theta$, sum to one, and are bounded
away from zero; Appendix.~\ref{app:timing} specifies the mapping. Interval $m$ has mean
speed $1/(M\Delta t_m)$ in path coordinates, so $\theta$ controls
relative traversal speeds while $\tau$ sets the overall duration.
Writing $\tau_0=0$ and $\tau_m=\sum_{j=1}^m\Delta t_j$, we construct
$s_{\theta,\tau}$ by monotone interpolation through the timing knots
$(\tau_m,m/M)$. Sampling the resulting motion gives
\begin{equation}
\begin{aligned}
\xi_{\theta,\tau}
&=\bigl((\gamma(s_{\theta,\tau}(t_i)),t_i)\bigr)_{i=1}^{n},\\
&\qquad 0=t_1<\cdots<t_n=\tau.
\end{aligned}
\label{eq:retimed-segment}
\end{equation}

We use the same trajectory cost $c_{\mathrm{traj}}(\xi_{\theta,\tau})$
to guide retiming, optimizing $\theta$ and $\tau$ using gradients
through the time scaling, fixed path, and frozen encoder. Candidate
motions are checked against the robot's motion limits before acceptance.
Appendix C gives the full objective and numerical optimization details.

\section{Experiments}\label{sec:experiments}
\begin{table*}[t]
\centering
\footnotesize
\setlength{\tabcolsep}{2.5pt}
\renewcommand{\arraystretch}{1.1}
\begin{tabular*}{\textwidth}{@{\extracolsep{\fill}}l*{12}{c}@{}}
\toprule
 & \multicolumn{4}{c}{\textbf{\taskplate{}}}
 & \multicolumn{4}{c}{\textbf{\taskpack{}}}
 & \multicolumn{4}{c}{\textbf{\tasksort{}}} \\
\cmidrule(lr){2-5}\cmidrule(lr){6-9}\cmidrule(lr){10-13}
 & \multicolumn{2}{c}{\emph{\evalmain{}}} & \multicolumn{2}{c}{\emph{\evalood{}}}
 & \multicolumn{2}{c}{\emph{\evalmain{}}} & \multicolumn{2}{c}{\emph{\evalood{}}}
 & \multicolumn{2}{c}{\emph{\evalmain{}}} & \multicolumn{2}{c}{\emph{\evalood{}}} \\
\cmidrule(lr){2-3}\cmidrule(lr){4-5}
\cmidrule(lr){6-7}\cmidrule(lr){8-9}
\cmidrule(lr){10-11}\cmidrule(lr){12-13}
Approach
 & Succ. & Prog. & Succ. & Prog.
 & Succ. & Prog. & Succ. & Prog.
 & Succ. & Prog. & Succ. & Prog. \\
\midrule
\basepretrained
 & 0\% & 18\% & \best{90\%} & \best{95\%}
 & 0\% & 13\% & \best{90\%} & \best{95\%}
 & 5\% & 50\% & \best{90\%} & \best{95\%} \\
\basetamp
 & 5\% & 35\% & 85\% & 93\%
 & 5\% & 37\% & 70\% & 85\%
 & 15\% & 58\% & 75\% & 88\% \\
\addlinespace[1.5pt]
\ablstyle
 & 10\% & 39\% & -- & --
 & 0\% & 31\% & -- & --
 & 35\% & 54\% & -- & -- \\
\abltiming
 & 0\% & 19\% & -- & --
 & 0\% & 18\% & -- & --
 & 0\% & 29\% & -- & -- \\
\abljoint
 & 35\% & 73\% & -- & --
 & 15\% & 53\% & -- & --
 & 20\% & 66\% & -- & -- \\
\ours
 & \best{65\%} & \best{84\%} & 80\% & 90\%
 & \best{35\%} & \best{81\%} & 85\% & 88\%
 & \best{70\%} & \best{89\%} & \best{90\%} & 93\% \\
\midrule
\baseteleop
 & 50\% & 93\% & 95\% & 98\%
 & 45\% & 90\% & 95\% & 98\%
 & 90\% & 98\% & 95\% & 98\% \\
\bottomrule
\end{tabular*}
\captionsetup{skip=4pt}
\caption{Success rate (Succ.) and task progress (Prog.) for task-specific
fine-tuning. Each task group reports performance on the named target task and
retention on \taskcloth{}; demonstrations of this task are excluded from
fine-tuning. \best{Bold} marks the best non-oracle result in each column. Human
teleoperation is an oracle only in its access to human target-task
demonstrations and is not assumed to upper-bound the autonomous methods.}
\label{tab:main-results}
\end{table*}

We evaluate whether distribution-aligned TAMP produces effective data for
fine-tuning a pretrained VLA. We address four questions:
(\textbf{Q1}) how does the demonstration source affect target-task performance;
(\textbf{Q2}) how much does each alignment component contribute;
(\textbf{Q3}) does task-specific fine-tuning retain the pretrained policy's
performance on \taskcloth{}; and (\textbf{Q4}) how does performance change over
the tested demonstration budgets?

\subsection{Experimental Setup} 
\begin{figure}[t]
    \centering
    \includegraphics[width=\linewidth]{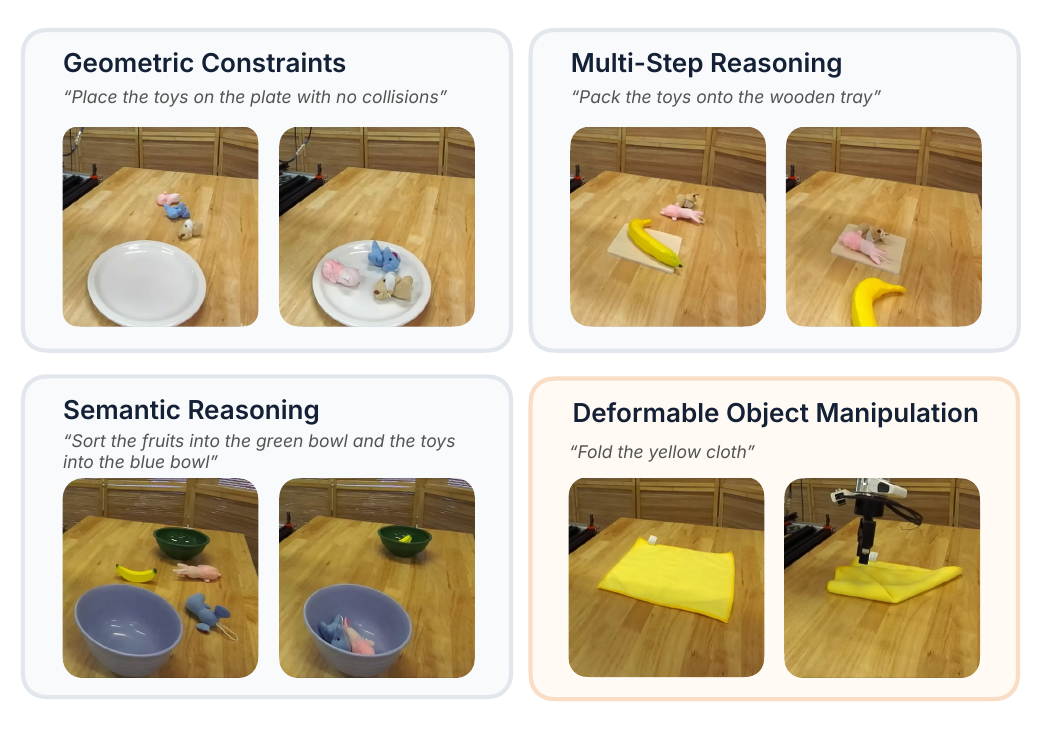}
    \hfill
    \caption{Task-specific fine-tuning and OOD evaluation. We fine-tune the same pretrained VLA using three different \ours{} demonstrations generated for each target task: \taskplate{}, \taskpack{}, and \tasksort{}. We evaluate each resulting task-specific policy on its corresponding target task and, without demonstrations of \taskcloth{}, on this task as an OOD evaluation of whether task-specific fine-tuning preserves pretrained capabilities.}
    \label{fig:tasks}
\end{figure}
\paragraph{Robot and Tasks}
We use the DROID robot platform~\cite{khazatsky2024droid}: a Franka Emika Panda
7-DOF arm with a Robotiq 2F-85 gripper, two external ZED 2
stereo cameras, and a wrist-mounted ZED Mini stereo camera. Human
demonstrations use Meta Quest 2 controllers. 
We provide an overview of the experimental tasks in Figure.~\ref{fig:tasks}. 
The three target tasks each require three object moves. \textbf{\taskplate{}} requires placing three toys
on a plate without collisions. Earlier placements constrain the remaining free
space; placing the first toy in the center can prevent a later placement.
\textbf{\taskpack{}} requires removing an obstruction from a tray before
placing two toys in it. \textbf{\tasksort{}} requires placing three objects
into two bowls according to their semantic categories. We test capability
retention on \textbf{\taskcloth{}}, which requires folding a cloth and lies
outside the capabilities of the planner. The pretrained policy can perform
this task, but the planner-generated data excludes it.

\paragraph{Approaches}
{\bfseries\boldmath\basepretrained{}}~\cite{black2025pi05} is the pretrained checkpoint
without task-specific fine-tuning. \textbf{\basetamp{}} denotes a VLA
fine-tuned on successful demonstrations generated by unmodified
TiPToP~\cite{shen2026tiptop}; its table entries measure the learned policy, not
the planner's execution success. \textbf{\ours{}} fine-tunes the same
pretrained checkpoint on demonstrations from our distribution-aligned
planner. The three ablations each remove one component while retaining the
other two: \textbf{\ablstyle{}}, \textbf{\abltiming{}}, and
\textbf{\abljoint{}}. Figure~\ref{fig:gap} compares the planner-generated data
sources with DROID in the joint-space and trajectory-latent representations
used by our objectives. \textbf{\baseteleop{}} is fine-tuned on
human-controlled target-task demonstrations. We refer to it as an oracle
because it has access to such demonstrations, not because it is an upper bound
on policy performance.

\paragraph{Evaluation Protocol and Metrics}
For every target task and fine-tuning approach, we train a separate checkpoint
from 20 successful demonstrations using the same fine-tuning procedure. The
demonstrations are not pooled across target tasks. \basepretrained{} uses no
target-task demonstrations. Each checkpoint is evaluated for 20 trials on
fresh initial scenes from the task's evaluation distribution; the reported
entries summarize those rollouts for one trained checkpoint. We also evaluate
each main checkpoint on \taskcloth{} after target-task fine-tuning. Human
assistance is permitted for scene setup and resets, but \ours{} plans and
executes the demonstration motions without human control.

We define binary, task-specific subgoals solely to score evaluation rollouts;
they are not training labels and are not provided to the policy. Each target
task has seven evaluation subgoals, and \taskcloth{} has two.
Appendix~\ref{app:evaluation-subgoals} gives their definitions. Each trial ends
when the policy completes the task or reaches the shared evaluation horizon. A
trial succeeds when the policy completes every subgoal. Task progress is the
fraction of subgoals completed at termination, normalized to 100\% and averaged
over the 20 trials. This score gives partial credit to rollouts that make
progress without completing the task.

\subsection{Experimental Results}

\begin{figure}[t]
    \centering
    \includegraphics[width=\linewidth]{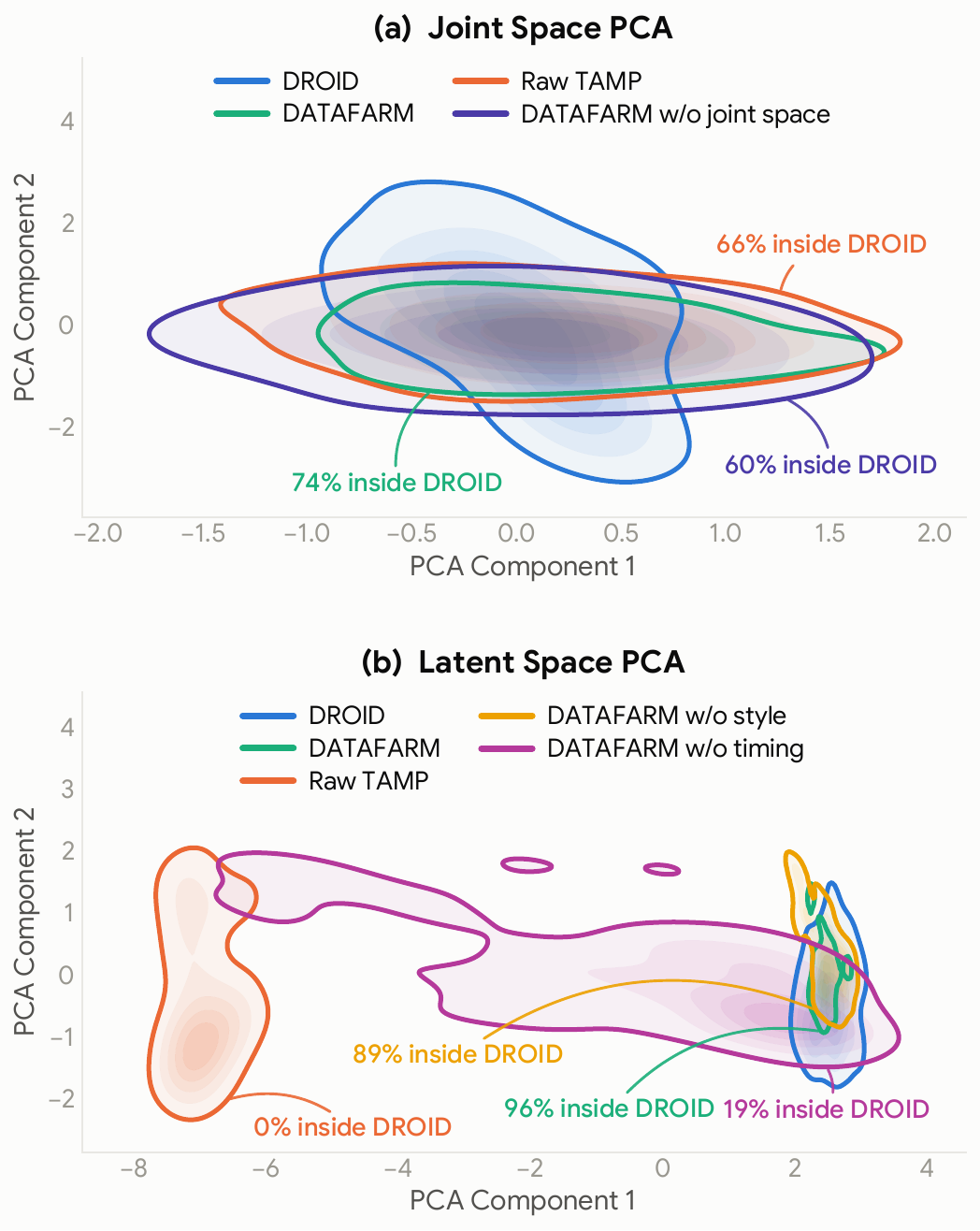}
    \hfill
    \caption{Alignment in two measured representations. Density contours show
    samples projected onto the first two principal components of (a) robot joint
    configurations and (b) trajectory embeddings. The outer line is the displayed
    density contour for each source. Each annotation reports the fraction of that
    source's projected samples inside the outer DROID contour. These projections
    diagnose overlap in the displayed representations; they do not measure equality
    with the full pretraining distribution.}
    \label{fig:gap}
\end{figure}

\paragraph{Effect of Demonstration Source}
Table~\ref{tab:main-results} compares the policies trained from each data
source. Averaged across the three target tasks, \basepretrained{} succeeds on
1.7\% of trials and fine-tuning on raw TAMP data raises success to 8.3\%, a
gain of 6.6 percentage points. The TiPToP demonstrations used for fine-tuning
are successful planner executions, so their limited downstream benefit cannot
be attributed to a failure to solve the tasks. A task-valid trajectory is not
necessarily effective fine-tuning data for this pretrained policy.

Fine-tuning on \ours{} demonstrations raises average success to 56.7\% and
average task progress to 84.7\%, compared with 43.3\% progress for raw TAMP and
27.0\% for \basepretrained{}. The improvement occurs on every target task:
\ours{} succeeds on 65\%, 35\%, and 70\% of trials, while raw TAMP succeeds on
5\%, 5\%, and 15\%, respectively. These results answer \textbf{Q1} because
changing how the planner generates demonstrations has a large effect even when
every training trajectory satisfies the task constraints.

\ours{} exceeds teleoperation on \taskplate{} (65\% versus 50\% success) and
trails it on \taskpack{} (35\% versus 45\%) and \tasksort{} (70\% versus 90\%).
Policies trained on human demonstrations achieve higher progress on all three
tasks. \ours{} requires human assistance only for scene setup and resets. Its
autonomously collected demonstrations yield 56.7\% average policy success,
nearly matching the 61.7\% obtained with human demonstrations.

\paragraph{Alignment with DROID}
\ours{} increases overlap with DROID in both the joint-space and
trajectory-latent projections (Figure~\ref{fig:gap}). In joint space, 74\% of
\ours{} samples lie inside the displayed DROID contour, compared with 66\%
for raw TAMP and 60\% without joint-space alignment. In trajectory latent
space, 96\% of \ours{} samples lie inside the DROID contour, compared with
0\% for raw TAMP, 89\% without style alignment, and 19\% without timing
alignment.

\paragraph{Contribution of Each Alignment Component}
We remove one alignment component at a time while retaining the other two
(Table~\ref{tab:main-results}). Removing joint-space alignment reduces average
target-task success from 56.7\% to 23.3\%, and removing style alignment reduces
it to 15.0\%. Removing timing alignment has the largest effect: the policy
fails every trial across all three target tasks. This ablation also reduces
projected latent overlap with DROID from 96\% to 19\%
(Figure~\ref{fig:gap}). The style ablation retains 89\% overlap but reaches
only 15.0\% success, so high overlap in this projection alone does not ensure
task success. Each component improves policy performance when added to the
other two (\textbf{Q2}).

\paragraph{Retention of Pretrained Capabilities}
Fine-tuning on \ours{} demonstrations largely preserves the pretrained
policy's performance on \taskcloth{} (\textbf{Q3}), a task that TiPToP cannot
perform. The unmodified
\basepretrained{} succeeds on 90\% of trials. After fine-tuning
on \taskplate{}, \taskpack{}, or \tasksort{}, the policy succeeds on 80\%,
85\%, and 90\% of \taskcloth{} trials, respectively
(Table~\ref{tab:main-results}). Although none of the fine-tuning datasets
include this task, the policies retain 85\% average success.

\begin{table}[t]
\centering
\small
\renewcommand{\arraystretch}{1.1}
\begin{tabular}{@{}lccccc@{}}
\toprule
 & \multicolumn{5}{c}{\textbf{\ours{} demonstrations}} \\
\cmidrule(lr){2-6}
 & 0 & 20 & 40 & 60 & 80 \\
\midrule
Success rate & 0\% & 65\% & 75\% & 75\% & 85\% \\
Task progress & 18\% & 84\% & 90\% & 94\% & 97\% \\
\bottomrule
\end{tabular}
\captionsetup{skip=4pt}
\caption{Policy performance on \taskplate{} with increasing numbers of
\ours{} demonstrations. The 0-demonstration column reports
\basepretrained{} without fine-tuning.}
\label{tab:scaling}
\end{table}

\paragraph{Effect of Demonstration Budget}
We evaluate how the demonstration budget affects policy performance on
\taskplate{} (\textbf{Q4}, Table~\ref{tab:scaling}). The pretrained checkpoint
starts at 0\% success and 18\% progress. Fine-tuning on 20 \ours{}
demonstrations raises success to 65\% and progress to 84\%. Success reaches
75\% with 40 demonstrations and remains at 75\% with 60. With 80
demonstrations, the policy reaches 85\% success and 97\% progress.

\section{Conclusion}\label{sec:conclusion}
Our experiments show that successful TAMP executions can still provide poor
training data for a pretrained VLA. Across three target tasks, fine-tuning
\basepretrained{} on raw TiPToP demonstrations raises average success from
1.7\% to only 8.3\%. \ours{} aligns planner-generated trajectories with DROID
in robot joint configurations, motion style, and execution timing. Policies
fine-tuned on these demonstrations achieve 56.7\% average success, nearly
matching the 61.7\% achieved with human demonstrations. They also retain 85\%
average success on \taskcloth{}, compared with 90\% for the pretrained policy,
even though TiPToP cannot perform this task and the fine-tuning data contain no
demonstrations of it. Removing any alignment component reduces target-task success;
removing timing alignment yields 0\% success. To our knowledge, \ours{} is the
first system to guide TAMP demonstration generation with a reference
distribution drawn from VLA pretraining data.

\paragraph{Limitations}
\ours{} inherits the task coverage and failure modes of its TAMP system. Our
implementation focuses on quasi-static pick-and-place manipulation. Extending
the planner to contact-rich and non-prehensile manipulation would allow
\ours{} to generate demonstrations for a broader set of tasks. The alignment
objectives require representative samples from the VLA's
pretraining distribution. We use DROID demonstrations~\cite{khazatsky2024droid}
as the reference. Applying \ours{} to a model whose pretraining data are
unavailable would require a suitable substitute dataset. Our evaluation covers
one pretrained VLA checkpoint, one robot embodiment, three target tasks, and
one retained capability. Experiments with other VLAs, embodiments, and task
families are needed to establish how broadly the three alignment components
improve fine-tuning.

\section{Acknowledgements}
We would like to thank Dhruv Shah, Mingtong Zhang, and Liang Ji for fruitful discussions. 
This work was partially supported by a Princeton SEAS Innovation, an NVIDIA Academic Grant Program, and a Princeton AI Lab Grant. 
We thank Google’s TPU Research Cloud (TRC) for providing access to Cloud TPUs.

\bibliographystyle{IEEEtran}
\bibliography{example,datafarm_related_work}

\newpage
\clearpage
\onecolumn

\setlength{\parskip}{1em}

\pagenumbering{arabic}%
\renewcommand*{\thepage}{A\arabic{page}} %
\appendix
This appendix defines the evaluation subgoals from
Sec.~\ref{sec:experiments} and provides additional implementation details for
Sec.~\ref{sec:approach}.
Appendix~\ref{app:evaluation-subgoals} gives the task-specific scoring criteria.
Appendix~\ref{app:latent} documents the trajectory latent space of Sec.~\ref{sec:reference-models}: the complete $113$-dimensional style fingerprint (Appendix~\ref{app:fingerprint}), the encoder architecture (Appendix~\ref{app:arch}), and training details (Appendix~\ref{app:encoder-training}).
Appendix~\ref{app:style} documents the style alignment objective of Sec.~\ref{sec:aligned-planning}, including the full gradient chain rule before the collapse to Eq.~\eqref{eq:style-grad}.
Appendix~\ref{app:timing} documents the timing alignment objective from the same section, and Appendix~\ref{app:joint} the joint-space alignment objective.

\startcontents[sections]
\printcontents[sections]{l}{1}{\setcounter{tocdepth}{2}}
\newpage
\subsection{Trajectory Latent Space}
\label{app:latent}

\subsubsection{Trajectory inputs}
\label{app:trajectory-inputs}
A timed segment $\xi=((q_i,t_i))_{i=1}^{n}$, with $t_1=0$, is resampled
at $F=15$\,Hz to configurations $\tilde q\in\mathbb R^{\tilde n\times7}$.
With $D$ the central-difference operator of step $1/F$, the encoder input stacks positions, velocities, accelerations,
and jerks:
\begin{equation}
Y=[\,Y_q\mid Y_v\mid Y_a\mid Y_j\,]
=[\,\tilde q\mid D\tilde q\mid D^2\tilde q\mid D^3\tilde q\,]
\in\mathbb R^{\tilde n\times28}.
\label{eq:features}
\end{equation}
Each channel is then standardized,
\begin{equation}
\bar Y=(Y-\mu_c)\oslash\sigma_c,
\label{eq:standardize}
\end{equation}
where $\mu_c,\sigma_c\in\mathbb R^{28}$ are fixed per-channel means and
standard deviations over the encoder's training dataset.

\subsubsection{Motion descriptors}
\label{app:fingerprint}

The auxiliary head $g_\psi$ predicts the $113$-dimensional descriptor vector $f(\xi)$ from a sampled latent vector. Table~\ref{tab:fingerprint-joint} lists the $14$ descriptors computed independently for each of the $7$ joints ($98$ entries). Table~\ref{tab:fingerprint-whole} lists the
$15$ whole-trajectory descriptors. We denote the speed as:
\begin{equation}
v(t)=\|\dot q(t)\|_2.
\label{eq:speed-signal}
\end{equation}
We write $\mathcal F\{\cdot\}$ for the discrete Fourier transform,
$S(\nu)$ for the normalized power spectrum of $v(t)$ at frequency
$\nu$, and $N_{\mathrm{freq}}$ for the number of frequency bins used
in the spectral sums. In the smoothness descriptor, $v_{\max}$ is
the maximum of $v(t)$. Spectral bands are low $[0,0.5)$\,Hz, mid
$[0.5,2)$\,Hz, and high $[2,7.5)$\,Hz, with the last bounded by
the Nyquist frequency at $F=15$\,Hz.

\begin{table}[ht]
\centering
\footnotesize
\setlength{\tabcolsep}{2.5pt}
\renewcommand{\arraystretch}{1.1}
\begin{tabular}{@{}p{0.26\linewidth} p{0.28\linewidth} p{0.40\linewidth}@{}}
\toprule
Descriptor & Description & Computation \\
\midrule
\multicolumn{3}{@{}l}{\textbf{Velocity}} \\
\texttt{vel\_mean} & mean joint speed & $\mathrm{mean}_t |\dot q^{(j)}|$ \\
\texttt{vel\_std} & variability of joint speed & $\mathrm{std}_t |\dot q^{(j)}|$ \\
\texttt{vel\_max} & peak joint speed & $\max_t |\dot q^{(j)}|$ \\
\texttt{vel\_p95} & near-peak speed, outlier-robust & $95$th pct.\ of $|\dot q^{(j)}|$ \\
\midrule
\multicolumn{3}{@{}l}{\textbf{Acceleration and jerk}} \\
\texttt{acc\_mean} & mean acceleration magnitude & $\mathrm{mean}_t |\ddot q^{(j)}|$ \\
\texttt{acc\_std} & variability of acceleration & $\mathrm{std}_t |\ddot q^{(j)}|$ \\
\texttt{acc\_max} & peak acceleration & $\max_t |\ddot q^{(j)}|$ \\
\texttt{jerk\_mean} & mean jerk magnitude & $\mathrm{mean}_t |\dddot q^{(j)}|$ \\
\texttt{jerk\_max} & peak jerk & $\max_t |\dddot q^{(j)}|$ \\
\midrule
\multicolumn{3}{@{}l}{\textbf{Extent and reversals}} \\
\texttt{rom} & range of motion & $\max_t q^{(j)} - \min_t q^{(j)}$ \\
\texttt{vsign} & direction-reversal rate & sign changes of $\dot q^{(j)}$, per $t_n$ \\
\midrule
\multicolumn{3}{@{}l}{\textbf{Spectral (share of velocity power per band)}} \\
\texttt{bandfrac\_low} & share that is slow, gross motion & frac.\ of $|\mathcal{F}\{\dot q^{(j)}\}|^2$ in $[0, 0.5)$\,Hz \\
\texttt{bandfrac\_mid} & share from corrections, wobble & frac.\ of $|\mathcal{F}\{\dot q^{(j)}\}|^2$ in $[0.5, 2)$\,Hz \\
\texttt{bandfrac\_high} & share from fast micro-motion & frac.\ of $|\mathcal{F}\{\dot q^{(j)}\}|^2$ in $[2, 7.5)$\,Hz \\
\bottomrule
\end{tabular}
\captionsetup{skip=4pt}
\caption{Per-joint motion descriptors. Fourteen descriptors for each of seven joints yield 98 entries.}
\label{tab:fingerprint-joint}
\end{table}

\begin{table}[ht]
\centering
\footnotesize
\setlength{\tabcolsep}{2.5pt}
\renewcommand{\arraystretch}{1.1}
\begin{tabular}{@{}p{0.26\linewidth} p{0.28\linewidth} p{0.40\linewidth}@{}}
\toprule
Descriptor & Description & Computation \\
\midrule
\multicolumn{3}{@{}l}{\textbf{Path geometry}} \\
\texttt{path\_len\_per\_s} & joint-space distance per second & $\big(\sum_{i=1}^{\tilde n-1}\|\tilde q_{i+1}-\tilde q_i\|_2\big)/t_n$ \\
\texttt{straightness} & how direct the path is & $\|\tilde q_{\tilde n}-\tilde q_1\|_2$ / path length \\
\midrule
\multicolumn{3}{@{}l}{\textbf{Speed}} \\
\texttt{speed\_mean} & mean overall speed & $\mathrm{mean}_t\, v(t)$ \\
\texttt{speed\_std} & speed variability & $\mathrm{std}_t\, v(t)$ \\
\texttt{speed\_max} & peak overall speed & $\max_t v(t)$ \\
\texttt{frac\_moving} & fraction of time in motion & frac.\ of frames with $v(t) > 0.02$\,rad/s \\
\midrule
\multicolumn{3}{@{}l}{\textbf{Smoothness}} \\
\texttt{sparc} & spectral arc length $\to 0$ is smoother \cite{balasubramanian2015sparc} & arc length of the normalized speed spectrum up to the $10$\,Hz cutoff (limited by $F/2$), truncated below amplitude $0.05$ \\
\texttt{ldlj} & log dimensionless jerk; higher is smoother \cite{hogan2009sensitivity} & $-\log \big( (t_n^3 / v_{\max}^2) \int \ddot{v}^2 \, dt \big)$ \\
\texttt{submov\_per\_sec} & submovement rate & peaks in $v(t)$ with prominence $\ge 0.05 \max_t v(t)$, per second \\
\midrule
\multicolumn{3}{@{}l}{\textbf{Spectral (shape of the speed spectrum)}} \\
\texttt{spec\_centroid} & where the spectrum is centered & $\sum_{\nu}\nu S(\nu)$ \\
\texttt{spec\_bandwidth} & how spread out the spectrum is & $\big( \sum_{\nu}(\nu-\text{centroid})^2S(\nu) \big)^{1/2}$ \\
\texttt{spec\_entropy} & how broadband the motion is & $-\sum_{\nu}S(\nu)\log S(\nu)\,/\,\log N_{\mathrm{freq}}$ \\
\texttt{bandfrac\_low\_speed} & slow share of the speed profile & frac.\ of $|\mathcal{F}\{v\}|^2$ in $[0, 0.5)$\,Hz \\
\texttt{bandfrac\_mid\_speed} & mid share of the speed profile & frac.\ of $|\mathcal{F}\{v\}|^2$ in $[0.5, 2)$\,Hz \\
\texttt{bandfrac\_high\_speed} & fast share of the speed profile & frac.\ of $|\mathcal{F}\{v\}|^2$ in $[2, 7.5)$\,Hz \\
\bottomrule
\end{tabular}
\captionsetup{skip=4pt}
\caption{Whole-trajectory motion descriptors (15 entries). $S(\nu)$ is the normalized power spectrum of the speed signal $v(t)$ in Eq.~\eqref{eq:speed-signal}.}
\label{tab:fingerprint-whole}
\end{table}

\subsubsection{Encoder architecture}
\label{app:arch}

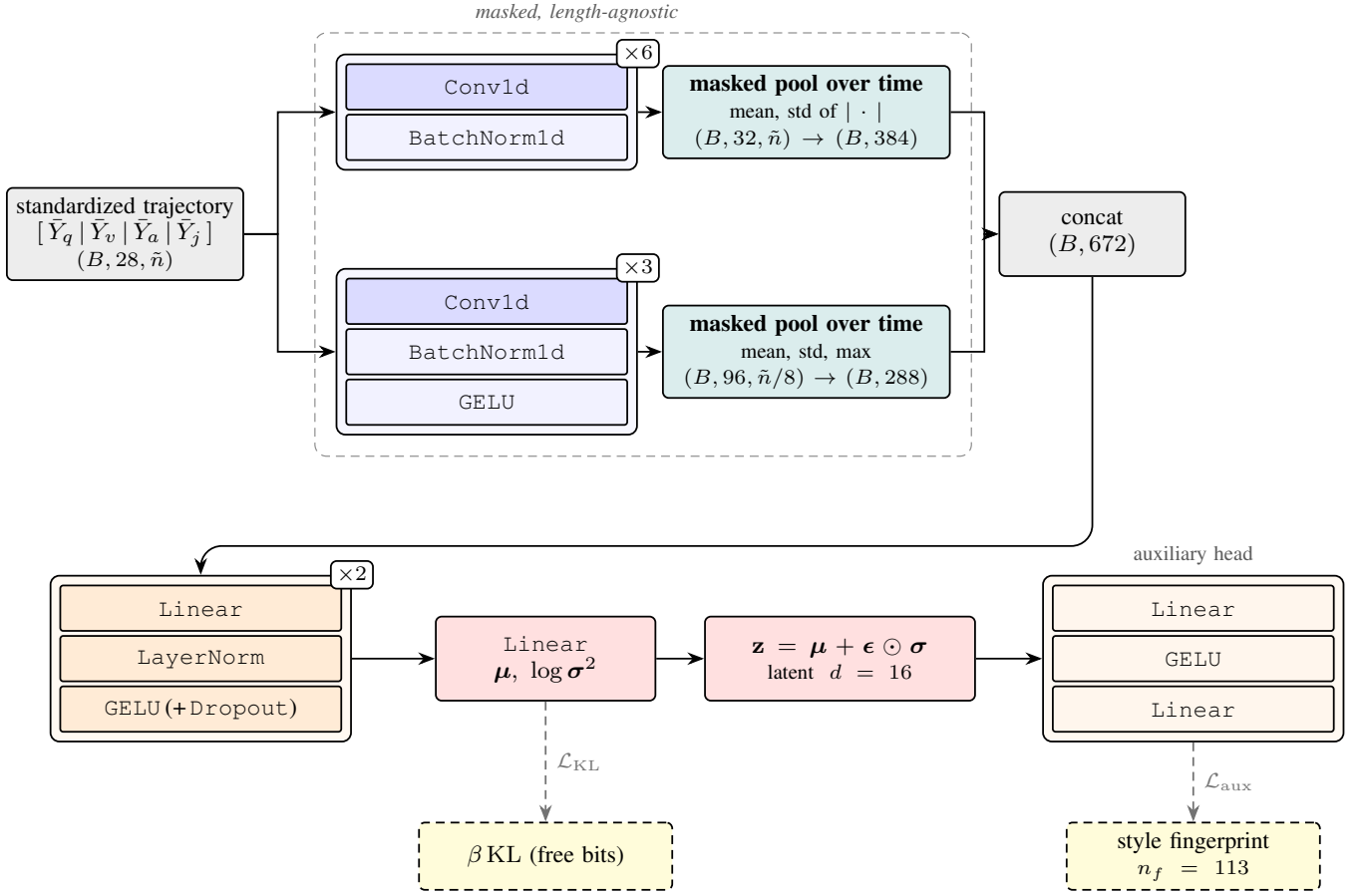
\begin{figure}[ht]
  \centering
  \resizebox{\linewidth}{!}{%
  \begin{tikzpicture}[
    font=\footnotesize,
    >={Stealth[length=2mm]},
    chip/.style={draw,semithick,rounded corners=1pt,align=center,
                 text width=32mm,minimum height=5.4mm,inner sep=2pt,fill=white},
    cconv/.style={chip,fill=blue!14},
    cnorm/.style={chip,fill=blue!5},
    cmlp/.style={chip,fill=orange!16},
    caux/.style={chip,fill=orange!7},
    box/.style={draw,semithick,rounded corners=2pt,align=center,
                text width=32mm,minimum height=10mm,inner sep=3pt},
    data/.style={box,fill=black!7,text width=26mm},
    pool/.style={box,fill=teal!14},
    lat/.style={box,fill=red!12,text width=24mm},
    side/.style={box,fill=yellow!18,densely dashed,text width=28mm,minimum height=8mm},
    grp/.style={draw,semithick,rounded corners=3pt,fill=blue!3,inner sep=3pt},
    badge/.style={draw,semithick,rounded corners=2pt,fill=white,
                  inner sep=2pt,font=\scriptsize\bfseries},
    note/.style={font=\scriptsize,text=black!65,align=center,text width=34mm},
    ar/.style={->,semithick},
    sar/.style={->,semithick,densely dashed,black!55},
  ]
  \hyphenpenalty=10000 \exhyphenpenalty=10000

  \node[data] (in) at (-8.5,1.1)
    {standardized trajectory\\$[\,\bar Y_q\,|\,\bar Y_v\,|\,\bar Y_a\,|\,\bar Y_j\,]$\\{\scriptsize$(B,28,\tilde n)$}};

  \node[cconv] (a1) at (-4.2,2.85) {\texttt{Conv1d}};
  \node[cnorm] (a2) at (-4.2,2.25) {\texttt{BatchNorm1d}};

  \node[cconv] (b1) at (-4.2,0.30) {\texttt{Conv1d}};
  \node[cnorm] (b2) at (-4.2,-0.30) {\texttt{BatchNorm1d}};
  \node[cnorm] (b3) at (-4.2,-0.90) {\texttt{GELU}};

  \node[pool] (pa) at (-0.4,2.55) {\textbf{masked pool over time}\\{\scriptsize mean, std of $|\cdot|$}\\{\scriptsize$(B,32,\tilde n)\to(B,384)$}};
  \node[pool] (pb) at (-0.4,-0.30) {\textbf{masked pool over time}\\{\scriptsize mean, std, max}\\{\scriptsize$(B,96,\tilde n/8)\to(B,288)$}};

  \node[box,fill=black!7,text width=20mm] (cat) at (3.0,1.1) {concat\\$(B, 672)$};

  \node[cmlp] (m1) at (-7.6,-3.35) {\texttt{Linear}};
  \node[cmlp] (m2) at (-7.6,-3.95) {\texttt{LayerNorm}};
  \node[cmlp] (m3) at (-7.6,-4.55) {\texttt{GELU}\,(+\,\texttt{Dropout})};

  \node[lat] (mu) at (-3.5,-3.95) {\texttt{Linear}\\$\boldsymbol{\mu},\ \log\boldsymbol{\sigma}^{2}$};
  \node[lat,text width=30mm] (z) at (0.0,-3.95)
    {$\mathbf{z}=\boldsymbol{\mu}+\boldsymbol{\epsilon}\odot\boldsymbol{\sigma}$\\{\scriptsize latent \ $d=16$}};

  \node[caux] (x1) at (4.2,-3.35) {\texttt{Linear}};
  \node[caux] (x2) at (4.2,-3.95) {\texttt{GELU}};
  \node[caux] (x3) at (4.2,-4.55) {\texttt{Linear}};

  \node[side] (kl) at (-3.5,-6.3) {$\beta\,\mathrm{KL}$ (free bits)};
  \node[side] (fp) at (4.2,-6.3) {style fingerprint\\{\scriptsize$n_f = 113$}};

  \begin{scope}[on background layer]
    \node[grp,fit=(a1)(a2)] (ga) {};
    \node[grp,fit=(b1)(b2)(b3)] (gb) {};
    \node[grp,fill=orange!5,fit=(m1)(m2)(m3)] (gm) {};
    \node[grp,fill=orange!5,fit=(x1)(x2)(x3)] (gx) {};
    \node[draw,densely dashed,black!45,rounded corners=4pt,inner sep=7pt,
          fit=(ga)(gb)(pa)(pb)] (enc) {};
  \end{scope}
  \node[badge] at (ga.north east) {$\times 6$};
  \node[badge] at (gb.north east) {$\times 3$};
  \node[badge] at (gm.north east) {$\times 2$};
  \node[note,anchor=south west,text width=60mm] at (enc.north west) {\itshape masked, length-agnostic};
  \node[note,anchor=south] at (gx.north) {auxiliary head};

  \draw[ar] (in.east) -- ++(4mm,0) |- (ga.west);
  \draw[ar] (in.east) -- ++(4mm,0) |- (gb.west);
  \draw[ar] (ga) -- (pa);
  \draw[ar] (gb) -- (pb);
  \draw[ar] (pa.east) -- ++(4mm,0) |- (cat.west);
  \draw[ar] (pb.east) -- ++(4mm,0) |- (cat.west);
  \draw[ar,rounded corners=2.5mm] (cat.south) -- ++(0,-3.2) -| (gm.north);
  \draw[ar] (gm) -- (mu);
  \draw[ar] (mu) -- (z);
  \draw[ar] (z)  -- (gx);
  \draw[sar] (mu) -- node[right,font=\scriptsize]{$\mathcal{L}_{\mathrm{KL}}$} (kl);
  \draw[sar] (gx) -- node[right,font=\scriptsize]{$\mathcal{L}_{\mathrm{aux}}$} (fp);
  \end{tikzpicture}}
  \vspace{6pt}
  \caption{Trajectory encoder architecture.}
  \label{fig:encoder-arch}
\end{figure}

Figure~\ref{fig:encoder-arch} shows the full trajectory encoder architecture.
It consists of two branches:
\paragraph{Spectral branch}
This branch uses a variety of kernel-dilation pairs to capture motion behaviors at different temporal scales. Their outputs are masked-pooled over
time by the mean and standard deviation of the absolute activations, giving a $384$-dimensional summary.

\paragraph{Temporal branch}
This branch captures the ordering of motion, such as acceleration during free-space transit followed by deceleration on approach to an object.

\subsubsection{Training details}
\label{app:encoder-training}

The trajectory encoder is trained in a self-supervised manner. For each trajectory, the objective
combines descriptor prediction with Gaussian KL regularization:
\begin{equation}
\begin{aligned}
\mathcal L(\phi,\psi;\xi)
={}&\mathbb E_\epsilon\!\left[
\left\|w\odot\bigl(g_\psi(z)-f(\xi)\bigr)\right\|_2^2\right]\\
&+\beta\,D_{\mathrm{KL}}\!\left(
\mathcal N\!\left(E_\phi(\xi),\operatorname{diag}(\sigma_\phi^2(\xi))\right)
\,\middle\|\,\mathcal N(0,I)\right),
\end{aligned}
\label{eq:encoder-training}
\end{equation}
where the encoder outputs a mean $E_\phi(\xi)\in\mathbb R^{d}$, with
$d=16$, and standard deviations $\sigma_\phi(\xi)$, and the latent is
sampled as
\begin{equation}
z=E_\phi(\xi)+\sigma_\phi(\xi)\odot\epsilon,
\qquad \epsilon\sim\mathcal N(0,I).
\label{eq:latent-sampling}
\end{equation}
Planning uses only the mean $E_\phi(\xi)$. The weights $w$
balance the different natural scales of the $113$ descriptors. We use
$\beta=0.005$ and apply free bits to the KL term to discourage latent
collapse while retaining information useful for descriptor prediction.

The encoder is trained on reference trajectories together with random
crops. Cropping increases the number of training examples and exposes
the encoder to partial motions of different lengths, matching the
segments queried during planning. Appendix~\ref{app:timing} explains
why these crops also motivate a boundary-speed penalty during retiming.
The channel statistics $(\mu_c,\sigma_c)$ are computed from the
reference inputs and kept fixed during training and planning. After
training, we freeze $\phi$ and fit the latent reference mean $\mu$
and covariance $\Sigma$ to embeddings of reference trajectories and
their crops. The inverse covariance $\Sigma^{-1}$ is used in
$c_{\mathrm{traj}}$ (Eq.~\eqref{eq:trajectory-cost}).
Table~\ref{tab:encoder-hparams} lists the training hyperparameters.

\begin{table}[ht]
\centering
\footnotesize
\setlength{\tabcolsep}{2.5pt}
\renewcommand{\arraystretch}{1.1}
\begin{tabular}{@{}p{0.52\linewidth} p{0.42\linewidth}@{}}
\toprule
Hyperparameter & Value \\
\midrule
\multicolumn{2}{@{}l}{\textbf{Data}} \\
Random crops per episode & $6$ \\
Crop length (fraction of episode) & $\mathcal U[0.4,\,1.0]$ \\
Encoder rate $F$ & $15$\,Hz \\
\midrule
\multicolumn{2}{@{}l}{\textbf{Model}} \\
Latent dimension $d$ & $16$ \\
Descriptor dimension & $113$ \\
Dropout & $0.3$ \\
\midrule
\multicolumn{2}{@{}l}{\textbf{Objective}} \\
KL weight $\beta$ & $0.005$, linear warmup over $10$ epochs \\
Free-bits threshold & $0.10$ nats per latent dimension \\
\midrule
\multicolumn{2}{@{}l}{\textbf{Optimization}} \\
Optimizer & Adam, weight decay $10^{-4}$ \\
Learning rate & $7\times10^{-4}$, cosine annealing \\
Batch size & $128$ \\
Epochs $\times$ steps per epoch & $60\times800$ \\
Gradient-norm clip & $5.0$ \\
\bottomrule
\end{tabular}
\captionsetup{skip=4pt}
\caption{Trajectory encoder training hyperparameters.}
\label{tab:encoder-hparams}
\end{table}

\subsection{Endpoint Selection}
\label{app:joint}

\subsubsection{Fitting the configuration density}

The configuration density $p_{\mathrm{conf}}$ is a $16$-component
full-covariance Gaussian mixture over the $7$ joint angles, fitted
to $987{,}631$ frames from $4{,}000$ reference episodes.
The configuration cost is $c_{\mathrm{conf}}(q)=-\log p_{\mathrm{conf}}(q)$.

\subsubsection{IK candidate generation and grasp symmetry}

For a candidate end-effector pose $x$, we obtain a finite set of IK
solutions $\mathcal Q(x)$ by searching from multiple initial guesses.
For grasps, we also consider the symmetry of the parallel-jaw gripper.
Let $R_\pi$ denote a rotation of $\pi$ about the gripper's local
approach axis. When the grasp model admits both $x$ and $xR_\pi$,
we solve IK for both representatives and select
\begin{equation}
q^\star=\underset{q\in\mathcal Q(x)\,\cup\,\mathcal Q(xR_\pi)}
{\operatorname{arg\,min}}\;c_{\mathrm{conf}}(q).
\label{eq:grasp-symmetry-selection}
\end{equation}
The stored grasp is updated to match the selected representative.
For poses without this symmetry, we rank only $\mathcal Q(x)$.
Candidates that fail a forward-kinematics check against their requested
pose are discarded. If none remain, we fall back to the planner's
baseline IK procedure. The associated grasps, placements, and connecting
motions remain subject to the planner's feasibility checks.

\subsubsection{Effect of including the rotated grasp}

IK branch selection for a fixed pose considers alternative arm
configurations but keeps the end-effector orientation fixed. Including
the rotated grasp also allows the planner to choose between the two
equivalent grasp orientations. Table~\ref{tab:twin} quantifies the effect
on $512$ endpoints from two scenes. We report the fraction of selected
configurations whose $c_{\mathrm{conf}}$ exceeds the $99$th percentile
of this cost on held-out reference data. Including the rotated grasp
reduces this fraction in both scenes.

\begin{table}[ht]
\centering
\footnotesize
\setlength{\tabcolsep}{2.5pt}
\renewcommand{\arraystretch}{1.1}
\begin{tabular}{@{}lcc@{}}
\toprule
Candidate pool & Semantic Reasoning & Geometric Constraints \\
\midrule
Best branch of the given pose only & $37$--$42\%$ & $37$--$42\%$ \\
Best branch over both grasp orientations & $5\%$ & $7\%$ \\
\bottomrule
\end{tabular}
\captionsetup{skip=4pt}
\caption{Share of $512$ selected endpoint configurations outside the $99$th percentile of the reference configuration density.}
\label{tab:twin}
\end{table}

\subsection{Path Optimization}
\label{app:style}

\subsubsection{Evaluation pipeline}

To evaluate $c_{\mathrm{traj}}(\xi)$, we resample the segment from the
trajectory optimizer's rate $1/\delta t$ to the encoder rate $F=15$\,Hz. We then form $Y$, standardize it, evaluate the mean embedding
$E_\phi(\xi)$, and compute the squared Mahalanobis cost in
Eq.~\eqref{eq:trajectory-cost}. The planner weights this cost by
$\lambda_{\mathrm{path}}$ in Eq.~\eqref{eq:aligned-trajopt}.

\subsubsection{Gradient computation and caching}
\label{app:style-grad}

Gradients propagate through the frozen encoder, channel
standardization, finite differences, and resampling. For the Jacobians
below, waypoint arrays and feature matrices are flattened in a fixed
order. Expanding the chain rule gives
\begin{equation}
\begin{aligned}
\frac{\partial c_{\mathrm{traj}}(\xi)}{\partial q_{1:n}}
={}&2\bigl(E_\phi(\xi)-\mu\bigr)^\top\Sigma^{-1}
\frac{\partial E_\phi(\xi)}{\partial\bar Y}\\
&\cdot\left[
\sum_{b\in\{q,v,a,j\}}
\frac{\partial\bar Y}{\partial Y_b}
\frac{\partial Y_b}{\partial\tilde q_{1:\tilde n}}
\right]
\frac{\partial\tilde q_{1:\tilde n}}{\partial q_{1:n}}.
\end{aligned}
\label{eq:style-grad-full}
\end{equation}
The derivative with respect to $\bar Y$ is through the neural network
following preprocessing. At fixed sample times, standardization,
finite differencing, and resampling form an affine map, so their
Jacobian is constant. We precompute
$J=\partial\bar Y/\partial q_{1:n}$ and reuse it across solver
iterations, obtaining
\begin{equation}
\frac{\partial c_{\mathrm{traj}}(\xi)}{\partial q_{1:n}}
=2\bigl(E_\phi(\xi)-\mu\bigr)^\top\Sigma^{-1}
\frac{\partial E_\phi(\xi)}{\partial\bar Y}\,J.
\label{eq:style-grad}
\end{equation}
Only the embedding's offset from the reference mean and the encoder
Jacobian vary with the current trajectory. A change in the sample
times or number of waypoints requires updating the preprocessing
operator. During path optimization, only the interior waypoint
configurations are updated; the endpoints remain fixed.

\subsection{Retiming}
\label{app:timing}

\subsubsection{Time-scaling implementation}

We use $M=64$ equal arc-length intervals in the time parameterization
of Eq.~\eqref{eq:time-parameterization}, with time fractions
\begin{equation}
\alpha_m(\theta)
=\frac{\exp(\tanh\theta_m)}{\sum_{j=1}^{M}\exp(\tanh\theta_j)}.
\label{eq:retiming-time-fractions}
\end{equation}
This applies an elementwise $\tanh$ to $\theta$ before softmax
normalization. Softmax alone can assign an arbitrarily small fraction
of the total duration to an interval, leading to near-zero traversal
times and large velocities or accelerations. Bounding the logits with
$\tanh$ limits the ratio between any two interval durations to at most
$e^2$, preventing individual time fractions from approaching zero.
This helps avoid allocations that violate the robot's motion limits,
but does not itself enforce velocity or acceleration limits, which are
checked separately.

The time scaling changes progress along the fixed path and preserves the
endpoint configurations, including those at gripper events.
The speed $\dot s_{\theta,\tau}(t)$ is expressed in normalized path
coordinates. The corresponding joint-space speed used in the boundary
penalty is
\begin{equation}
v(t)=\left\|\gamma'\bigl(s_{\theta,\tau}(t)\bigr)\right\|_2
\dot s_{\theta,\tau}(t).
\label{eq:retiming-joint-speed}
\end{equation}

\subsubsection{Boundary-speed penalty}

Let $v_{\mathrm{start}}(\xi)$ and $v_{\mathrm{end}}(\xi)$ denote the
joint-space speeds near a segment's boundaries, and let
$\bar v_{\mathrm{grip}}$ be the mean speed measured immediately before
and after gripper events in the reference demonstrations. We define
\begin{equation}
\begin{aligned}
c_{\mathrm{boundary}}(\xi)
={}&\bigl(v_{\mathrm{start}}(\xi)-\bar v_{\mathrm{grip}}\bigr)^2\\
&+\bigl(v_{\mathrm{end}}(\xi)-\bar v_{\mathrm{grip}}\bigr)^2.
\end{aligned}
\label{eq:boundary-cost}
\end{equation}
The full retiming objective combines the trajectory cost with this
penalty, weighted by $\lambda_{\mathrm{boundary}}$:
\begin{equation}
\min_{\theta,\,\tau>0}\quad
c_{\mathrm{traj}}(\xi_{\theta,\tau})
+\lambda_{\mathrm{boundary}}c_{\mathrm{boundary}}(\xi_{\theta,\tau}).
\label{eq:aligned-retiming}
\end{equation}

The reference density is fitted to embeddings of trajectories and random crops. In these windows, the median joint speed at the first and last frames is $0.296$\,rad/s, compared with $0.335$\,rad/s in their interiors. Thus, the latent cost alone does not reliably encourage slow motion at segment boundaries. Without the boundary term, the retiming optimizer produces boundary speeds of $0.30$--$0.37$\,rad/s. The reference target $\bar v_{\mathrm{grip}}=0.07$\,rad/s is measured by averaging joint speeds at frames immediately before and after gripper events in reference episodes.

\subsubsection{Optimization and candidate selection}

We optimize Eq.~\eqref{eq:aligned-retiming} using
Adam~\cite{kingma2015adam} from $16$ initial durations in parallel,
using gradients through the
path interpolation and frozen encoder. Each candidate is resampled
at the robot's control timestep and rejected if it exceeds velocity
or acceleration limits. Each survivor is then resampled at exactly
$F=15$\,Hz and evaluated with $c_{\mathrm{traj}}$; the candidate with
the lowest cost is returned. The boundary penalty therefore contributes
to optimization, while the final ranking uses the latent trajectory cost.

\subsection{Planner and Fine-Tuning Hyperparameters}
\label{app:hparams}

Table~\ref{tab:planner-hparams} lists the planner hyperparameters used
for data generation, and Table~\ref{tab:vla-hparams} lists the VLA
fine-tuning hyperparameters. All other planner settings use the
defaults of the base TAMP system.

\begin{table}[ht]
\centering
\footnotesize
\setlength{\tabcolsep}{2.5pt}
\renewcommand{\arraystretch}{1.1}
\begin{tabular}{@{}p{0.58\linewidth} p{0.36\linewidth}@{}}
\toprule
Hyperparameter & Value \\
\midrule
\multicolumn{2}{@{}l}{\textbf{Endpoint selection}} \\
IK solutions per candidate pose ($|\mathcal Q(x)|$) & $12$ \\
\midrule
\multicolumn{2}{@{}l}{\textbf{Path optimization}} \\
Trajectory-cost weight ($\lambda_{\mathrm{path}}$) & $25{,}000$ \\
\midrule
\multicolumn{2}{@{}l}{\textbf{Retiming}} \\
Arc-length intervals ($M$) & $64$ \\
Parallel initial durations & $16$ \\
Adam iterations & $500$ \\
Learning rate, time-allocation, ($\theta$) & $0.08$ \\
Learning rate, total duration ($\tau$) & $0.0375$ \\
Boundary-penalty weight ($\lambda_{\mathrm{boundary}}$) & $50$ \\
Gripper-event speed target ($\bar v_{\mathrm{grip}}$) & $0.07$\,rad/s \\
\bottomrule
\end{tabular}
\captionsetup{skip=4pt}
\caption{Planner hyperparameters used for data generation.}
\label{tab:planner-hparams}
\end{table}

\begin{table}[ht]
\centering
\footnotesize
\setlength{\tabcolsep}{2.5pt}
\renewcommand{\arraystretch}{1.1}
\begin{tabular}{@{}p{0.52\linewidth} p{0.42\linewidth}@{}}
\toprule
Hyperparameter & Value \\
\midrule
Base policy & $\pi_{0.5}$-DROID \\
Action horizon & $16$ \\
Optimizer & AdamW ($\beta_1{=}0.9$, $\beta_2{=}0.95$) \\
Peak learning rate & $2.5\times10^{-5}$ \\
Learning-rate schedule & $1{,}000$ warmup steps, cosine decay to $2.5\times10^{-6}$\\
Gradient-norm clip & $1.0$ \\
EMA decay & $0.99$ \\
Batch size & $256$ \\
Training steps & $20{,}000$ \\
\bottomrule
\end{tabular}
\captionsetup{skip=4pt}
\caption{VLA fine-tuning hyperparameters, shared by all data sources.}
\label{tab:vla-hparams}
\end{table}

\subsection{Task Evaluation Subgoals}
\label{app:evaluation-subgoals}

We use the binary subgoals in Table~\ref{tab:eval-rubrics} to score evaluation rollouts. A rollout succeeds when it meets every subgoal for its task, and task progress is the fraction of subgoals met at termination. We report the mean progress and the success rate over rollouts.

\begin{table}[ht]
\centering
\footnotesize
\setlength{\tabcolsep}{2.5pt}
\renewcommand{\arraystretch}{1.1}
\begin{tabular}{@{}p{0.30\linewidth} p{0.64\linewidth}@{}}
\toprule
Subgoal & Criterion \\
\midrule
\multicolumn{2}{@{}l}{\textbf{\taskplate{} ($7$ subgoals)}} \\
\texttt{blue\_toy\_on\_plate} & Blue toy ends up on the plate \\
\texttt{brown\_toy\_on\_plate} & Brown toy ends up on the plate \\
\texttt{pink\_toy\_on\_plate} & Pink toy ends up on the plate \\
\texttt{blue\_toy\_lifted} & Blue toy is lifted off the table at some point \\
\texttt{brown\_toy\_lifted} & Brown toy is lifted off the table at some point \\
\texttt{pink\_toy\_lifted} & Pink toy is lifted off the table at some point \\
\texttt{items\_spaced} & At least two toys are on the plate, and no placed toys touch \\
\midrule
\multicolumn{2}{@{}l}{\textbf{\tasksort{} ($7$ subgoals)}} \\
\texttt{blue\_toy\_in\_bowl} & Blue elephant plush ends up in the toy bowl \\
\texttt{pink\_toy\_in\_bowl} & Pink plush ends up in the toy bowl \\
\texttt{banana\_in\_bowl} & Banana ends up in the food bowl \\
\texttt{blue\_toy\_lifted} & Blue elephant plush is lifted off the table at some point \\
\texttt{pink\_toy\_lifted} & Pink plush is lifted off the table at some point \\
\texttt{banana\_lifted} & Banana is lifted off the table at some point \\
\texttt{sort\_clean} & At least two objects are placed, and neither bowl holds both a toy and a food item \\
\midrule
\multicolumn{2}{@{}l}{\textbf{\taskpack{} ($7$ subgoals)}} \\
\texttt{banana\_off\_tray} & Banana ends up released on the table, off the tray \\
\texttt{pink\_toy\_on\_tray} & Pink plush ends up resting on the tray itself \\
\texttt{brown\_toy\_on\_tray} & Brown plush ends up resting on the tray itself \\
\texttt{banana\_lifted} & Banana is lifted off the tray at some point \\
\texttt{pink\_toy\_lifted} & Pink plush is lifted off the table at some point \\
\texttt{brown\_toy\_lifted} & Brown plush is lifted off the table at some point \\
\texttt{tray\_cleared\_first} & At least one toy is placed on the tray, and the banana is already off the tray when the first toy is placed \\
\midrule
\multicolumn{2}{@{}l}{\textbf{\taskcloth{} ($2$ subgoals)}} \\
\texttt{fold\_any} & The cloth is folded any amount \\
\texttt{fold\_large} & The folded cloth covers less than $60\%$ of its original area on the table \\
\bottomrule
\end{tabular}
\captionsetup{skip=4pt}
\caption{Task-specific evaluation subgoals and success criteria.}
\label{tab:eval-rubrics}
\end{table}

\subsection{Failure Breakdown}
\label{app:failure-breakdown}

\begin{table*}[t]
\centering
\footnotesize
\setlength{\tabcolsep}{2.5pt}
\renewcommand{\arraystretch}{1.1}
\begin{tabular*}{\textwidth}{@{\extracolsep{\fill}}l*{7}{c}@{}}
\toprule
Approach & Number of Failures & Grasping & \begin{tabular}[b]{@{}c@{}}Manipulation/\\Placement\end{tabular} & \begin{tabular}[b]{@{}c@{}}Joint-Space\\Misalignment\end{tabular} & Semantic & \begin{tabular}[b]{@{}c@{}}Timeout\\(Stuck)\end{tabular} & \begin{tabular}[b]{@{}c@{}}Timeout\\(Did Not Finish)\end{tabular} \\
\midrule
\multicolumn{8}{@{}l}{\textbf{\taskplate{}}} \\
\basepretrained & 20 & 10\% & \underline{30\%} & 15\% & \textbf{45\%} & 0\% & 0\% \\
\basetamp & 19 & \underline{42\%} & \textbf{47\%} & 5\% & 5\% & 0\% & 0\% \\
\addlinespace[1.5pt]
\ablstyle & 18 & \textbf{89\%} & \underline{6\%} & \underline{6\%} & 0\% & 0\% & 0\% \\
\abltiming & 20 & 15\% & 10\% & 5\% & 5\% & \underline{20\%} & \textbf{45\%} \\
\abljoint & 13 & 0\% & \textbf{46\%} & \textbf{46\%} & 0\% & 0\% & \underline{8\%} \\
\ours & 7 & \textbf{71\%} & \underline{29\%} & 0\% & 0\% & 0\% & 0\% \\
\addlinespace[1.5pt]
\baseteleop & 10 & 0\% & \textbf{100\%} & 0\% & 0\% & 0\% & 0\% \\
\midrule
\multicolumn{8}{@{}l}{\textbf{\taskpack{}}} \\
\basepretrained & 20 & 0\% & 0\% & \underline{15\%} & \textbf{85\%} & 0\% & 0\% \\
\basetamp & 19 & \underline{37\%} & \textbf{42\%} & 5\% & 0\% & 0\% & 16\% \\
\addlinespace[1.5pt]
\ablstyle & 20 & \textbf{70\%} & \underline{20\%} & 0\% & 10\% & 0\% & 0\% \\
\abltiming & 20 & 15\% & 10\% & 0\% & 5\% & \underline{20\%} & \textbf{50\%} \\
\abljoint & 17 & \textbf{35\%} & \underline{18\%} & \textbf{35\%} & 6\% & 0\% & 6\% \\
\ours & 13 & \underline{15\%} & \textbf{69\%} & 0\% & \underline{15\%} & 0\% & 0\% \\
\addlinespace[1.5pt]
\baseteleop & 11 & 0\% & \textbf{82\%} & 0\% & \underline{18\%} & 0\% & 0\% \\
\midrule
\multicolumn{8}{@{}l}{\textbf{\tasksort{}}} \\
\basepretrained & 19 & 0\% & 0\% & \underline{11\%} & \textbf{89\%} & 0\% & 0\% \\
\basetamp & 17 & 6\% & \underline{18\%} & 12\% & \textbf{53\%} & 0\% & 12\% \\
\addlinespace[1.5pt]
\ablstyle & 13 & \textbf{62\%} & 8\% & 0\% & \underline{23\%} & 0\% & 8\% \\
\abltiming & 20 & 10\% & 5\% & 0\% & \underline{20\%} & 10\% & \textbf{55\%} \\
\abljoint & 16 & \underline{19\%} & 0\% & \textbf{56\%} & 6\% & 0\% & \underline{19\%} \\
\ours & 6 & \textbf{50\%} & 0\% & 0\% & \textbf{50\%} & 0\% & 0\% \\
\addlinespace[1.5pt]
\baseteleop & 2 & 0\% & \textbf{50\%} & 0\% & \textbf{50\%} & 0\% & 0\% \\
\bottomrule
\end{tabular*}
\captionsetup{skip=4pt}
\caption{Failure modes among failed rollouts. The count gives the number of
failures for each approach and task. \textbf{Bold} marks the most common mode
in each row; \underline{underline} marks the runner-up.}
\label{tab:failure-analysis}
\end{table*}

Table~\ref{tab:failure-analysis} breaks down the failed rollouts of each policy by failure mode. \emph{Grasping} failures miss or drop an object during pickup. \emph{Manipulation/placement} failures grasp the object
but release it in the wrong place or disturb other objects. \emph{Joint-space misalignment} failures move the arm into unusual postures from which it cannot recover or complete the task. \emph{Semantic} failures act on the wrong object
or goal, or ignore the instruction. The two \emph{timeout} modes separate policies that stop making progress in the middle of the episode (\emph{stuck}) from policies that are still making progress when the evaluation horizon ends (\emph{did not finish}).

\end{document}